\documentclass[10pt,twocolumn]{article}
\usepackage{graphicx}
\usepackage{amsmath,amssymb,pifont,times,cite}
\newcommand{\bm}[1]{\mathbf{#1}} %Bold vectors and matrices

\newcommand\raiseT[2]{%
\setbox0\hbox{$#1{#2}$}\raise\dp0\box0}

\usepackage{algorithm,algorithmic}
\makeatletter
\newcommand{\ALOOP}[1]{\ALC@it\algorithmicloop\ #1%
  \begin{ALC@loop}}
\newcommand{\ENDALOOP}{\end{ALC@loop}\ALC@it\algorithmicendloop}

\makeatother
\usepackage{comment}
\usepackage{url}
\usepackage{booktabs}
\usepackage[usenames,dvipsnames,table]{xcolor}

\definecolor{lightgray}{gray}{0.9}
\definecolor{lightblue}{rgb}{0.93,0.95,1.0}
\definecolor{aqua}{rgb}{0.0, 1.0, 1.0}
\usepackage{multirow}
\usepackage{verbatim}
\usepackage{hyperref} 
\hypersetup{ 
colorlinks = true, 
linkcolor={blue!90!black}, 
citecolor={blue!90!black}, 
urlcolor={blue!90!black} 
}

\title{\Large\textbf{Knowledge Distillation approach towards Melanoma Detection}}
\author{Md. Shakib Khan{{\fontsize{12}{12}\selectfont\textsuperscript{1,*}}}, Kazi Nabiul Alam{{\fontsize{12}{12}\selectfont\textsuperscript{1}}}, Abdur Rab Dhruba{{\fontsize{12}{12}\selectfont\textsuperscript{1}}}, Hasib Zunair{{\fontsize{12}{12}\selectfont\textsuperscript{2}}}, Nabeel Mohammed{{\fontsize{12}{12}\selectfont\textsuperscript{1}}} \\
{{\fontsize{12}{12}\selectfont\textsuperscript{1}}}North South University, Dhaka, Bangladesh \\
{{\fontsize{12}{12}\selectfont\textsuperscript{2}}}Concordia University, Montreal, QC, Canada \\
{{\fontsize{12}{12}\selectfont\textsuperscript{*}}}Corresponding Author: Md. Shakib Khan, E-mail: shakib.khan17@northsouth.edu
}
\date{}

\begin{document}
\maketitle

\begin{abstract}

Melanoma is regarded as the most threatening among all skin cancers. There is a pressing need to build systems which can aid in the early detection of melanoma and enable timely treatment to patients. Recent methods are geared towards machine learning based systems where the task is posed as image recognition, tag dermoscopic images of skin lesions as melanoma or non-melanoma. Even though these methods show promising results in terms of accuracy, they are computationally quite expensive to train, that questions the ability of these models to be deployable in a clinical setting or memory constraint devices. To address this issue, we focus on building simple and performant models having few layers, less than ten compared to hundreds. As well as with fewer learnable parameters, ~0.26 million (M) compared to 42.5M using knowledge distillation with the goal to detect melanoma from dermoscopic images. First, we train a teacher model using a ResNet-50 to detect melanoma. Using the teacher model, we train the student model known as Distilled Student Network (DSNet) which has around 0.26M parameters using knowledge distillation achieving an accuracy of 91.7\%. We compare against ImageNet pre-trained models such MobileNet, VGG-16, Inception-V3, EfficientNet-B0, ResNet-50 and ResNet-101. We find that our approach works well in terms of inference runtime compared to other pre-trained models, 2.57 seconds compared to 14.55 seconds. We find that DSNet (0.26M parameters), which is  15 times smaller, consistently performs better than EfficientNet-B0 (4M parameters) in both melanoma and non-melanoma detection across Precision, Recall and F1 scores.
\end{abstract}


\section{Introduction}
Skin is considered the largest organ in the human body that contains different layers and skin cancer forms on the surface which is also known as the epidermis area. Among different types of skin cancer, melanoma is known to be the deadliest among other kinds such as basal and squamous carcinoma. Melanoma generally forms in the pigment generating cells called melanocytes, which produce melanin and give different dark colors through these pigments in the epidermis area. Statistics from the World Health Organization (WHO) show that there are thousands of cases all around the globe with a vulnerable number of death cases in the year 2020. According to that, there are 324,635 new cases all around the globe, of which a total of 57,043 are death cases. They also show that 18 people out of 100, can’t survive who have been diagnosed with melanoma~\cite{r1}. There is no doubt this is an alarming matter and of great concern to dermatologists and health researchers all across the globe. People are mostly affected by such skin cancers, mostly Melanoma, due to ultraviolet rays from the sun. Sometimes genetic or other factors may occur, but these carcinomas can be beaten and cured if detected at an early stage of diagnosis~\cite{r2},~\cite{r3},~\cite{r4}.

Dermatologists use various techniques to identify the malignancy of skin lesions. Screening processes can be automated or predicted by symptoms. Biopsy is a well-known screening process that is used to detect if the growth of a skin lesion is malignant or benign. There are several types of biopsy processes, such as needle biopsy, computed tomography (CT)-guided biopsy, ultrasound-guided biopsy, bone \& bone marrow biopsy, etc. This is a clinical process of malignancy detection as it undergoes several pathological steps in the laboratory. Needle biopsy is done at a site where the needle is inserted with a small risk of bleeding and infection. Before needle biopsy, patients may also undergo visualization tests, such as a CT scan or an ultrasound. These tests are occasionally used during needle biopsy to identify the region to be biopsied more appropriately. After a needle biopsy, some mild pain can be anticipated, though it is usually controlled with pain relieving medications. But this whole procedure is stressful for patients mainly due to its invasive nature~\cite{r5}. Several studies have been conducted on mental distress related to the biopsy process showing the negative effects. The whole needling procedure of biopsy gives fear to many, especially patients who are already mentally upset about the malignancy result. Besides, the whole waiting time~\cite{r6} from providing a sample to getting back diagnosis results, is more stressful and affects their quality of life which creates disability to respond properly to the disease as well as decreases their ability to maintain a stable attitude towards family, work and overall life~\cite{r7}. Dermoscopy~\cite{r8} is a noninvasive method of examining pigmented, anatomical architectures of the epidermis, dermoepidermal junction and superficial, non-visible dermis with the naked eye. These tools, ranging from widely used, cheap hand-held instrumentation such as: Dermatoscope or Episcope with ten times higher fixed magnification, binocular microscopes, digital video microscopy, such tools have been used in dermoscopy procedures and these last 2 options provide higher magnification. Results showed that this method has more diverse behavior than naked-eye melanoma detection dermatology specialists working in this field and experienced in dermatology. Dermoscopy is consequently more successful than clinical evaluation in diagnosing melanoma in a pigmented skin lesion. In the last few years, with the improvement of Artificial Intelligence (AI) and in particular deep learning ~\cite{r9}, there have been several studies on deep learning based tools ~\cite{r5*} for computer aided medical diagnosis. Patients of this generation sometimes have faith issues or e and cause a lack of interest in the pathological screening process, resulting in AI-driven tools coming into the spotlight ~\cite{r10}. Skin lesion images can also be detected in an automatic way by framing the problem as image classification which is a rudimentary issue in computer vision. Basically, given a skin lesion image, the goal is to assign a label (e.g., malignant, benign) to that image. The classification methods are built on different deep learning techniques blended with computer vision and image processing methods. Transfer learning, Convolution Neural Networks (CNN), Deep Belief Networks (DBN), Stacked Sparse Autoencoder (SSAE), mixed hybrid methods, such deep learning techniques are broadly implemented in skin lesion classification~\cite{r11}. Medical images are different from natural images, where medical images contain more specific details of the lesion. We can demonstrate our optical or open-eye findings from natural images, where medical images are more fine-grained. It means that medical images occupy a very small portion of the entire image such as cardiomegaly in chest X-rays~\cite{r12} and microaneurysms in retinal images~\cite{r13}. X-rays ~\cite{r6*}, ultrasound scans, and other devices can be used to collect medical images. These images are most often used by physicians and medical professionals to evaluate or more closely monitor medical conditions, concerns, or constraints. Like microaneurysms, which is a small saccular outpouching that occurs due to several abnormal conditions like diabetic retinopathy. It is very tiny, even less than 125µm in diameter and the MA pixels account for less than 0.5\% of the retinal image~\cite{r13}. Another difference is the viewing point of medical images. The posteroanterior~\cite{r12} (PA) view available for any medical images. Sometimes the PA perspective is insufficient to make a diagnosis. Other perspectives, such as the lateral (L) view, are only relevant for certain diagnosis. In certain situations, the lateral view offers diagnostic information that is not obvious or accessible on the PA view.

Few works have been done in the direction of KD in medical image analysis tasks (e.g. CT, X-ray and ocular images)~\cite{r1*}~\cite{r2*}~\cite{r3*}~\cite{r4*}. To the best of our knowledge, there is no work on KD in dermatology images for skin cancer detection tasks, which is a core contribution of our work. We hope that this study would encourage the medical community as well computer vision community to focus on KD in dermatology image analysis tasks.

In this paper, we propose a knowledge distillation  based approach towards melanoma detection. The goal is build a model with few parameters that can accurately detect melanoma and can be easily integrated without requiring much compute. It enables us to build small and performant models which can be easily deployed in clinical settings without the need for heavy computational costs. First, we train a teacher model, then we train a student model (DSNet), which has a few thousand parameters that uses the teacher to distill knowledge. The main contributions of this paper can be summarized as follows:
\begin{itemize}

\item We propose the first knowledge distillation (KD) approach in detecting melanoma from dermoscopic images of skin lesions with the goal of reducing model complexity and making deployment more feasible.

\item We show that our approach in comparison to other pre-trained model requires significantly smaller inference time, 2.57 seconds compared to 14.55 seconds.

\item We show that our approach not only performs better than EfficientNet-B0 in term of inference runtime but also in terms of accuracy for both melanoma and non-melanoma classes across multiple metrics. 
\end{itemize}

The remaining parts of this paper are organized as follows. In Section \ref{RelatedWorks}, we discussed the related study on medical image classification, specifically melanoma detection and tasks done with the Knowledge Distillation approach. Section \ref{Method} demonstrates the methods and tools used in our work to classify melanoma with a knowledge distillation approach. In Section \ref{experiments}, experiments are performed on dermoscopic images from datasets, to find out the effectiveness of the proposed approach in comparison with baseline methods and we discussed our contribution and findings. Finally, concluded in Section \ref{conclusion}.

\section{Related Works} \label{RelatedWorks}
Several researches have been initiated to see if machine learning can be used to diagnose skin cancer merely from pictures of skin lesions. Authors in~\cite{r14}, proposed a ‘Combined Convolutional Neural Network’ (cCNN) method. They trained the network with 7895 dermoscopic images with a CNN and 5829 close-up images of lesions with another CNN. They combined the output of two CNNs by XGBoost and then they compared their results with human raters. They showed that the cCNN model predicted the same as that of a group of specialists in dermatology, and they also claimed that improved unnoisy data can give more efficiency that can give better outcomes compared to specialists' predictions. In~\cite{r15}, the authors developed a multi-stage framework regarding multiple neural networks linked to couple of classification levels. The first level includes five classifications: the perceptron coupled with a local color binary pattern, the perceptron and oriented gradient color histograms, the generative adversary (segmentation) network coupled with the ABCD rule, the ResNet~\cite{r16}, and the AlexNet~\cite{r9}. Here, the ABCD block was used to perform the 4 different features and to figure out the classification outcome. The ABCD formula consists of: Asymmetry (As), border score (B), Color variation score (C), and Diameter (D). And another level performs backpropagation perceptron to get the final output (melanoma or non-melanoma), and finally these steps end up with 97.5\% accuracy [15]. Researchers improved their model by implementing FRCN-88~\cite{r17} to avoid the vanishing gradient problem, and they constructed a new framework Lesion Indexing Network (LIN)~\cite{r18} built with the MatConvNet toolbox for skin lesion image classification on the ISIC 2017~\cite{r19} dataset. Then they compared AlexNet, VGG-16, ResNet-50/100 with their LIN network which achieved better results. In~\cite{r20}, authors investigate the performance of Deep Convolutional Neural Network (DCNNs) to determine skin lesions into 7 particular categories: nevus melanocytic, melanoma, benign keratosis, carcinoma basal cell, actinic keratosis, vascular lesions and dermatofibroma. This work aimed at assessing the efficiency of the EfficientNet~\cite{r21} models. Two advanced architectures of computer vision, such as ResNet50 and the EfficientNet-B0, were used to perform the classification technique on the HAM10000~\cite{r22} dataset. In their study, the EfficientNet-B0 model was found to outperform the ResNet-50 model with fewer parameters for each classification category. EfficientNet-B0 produced better AUC ROCs, and also achieved higher average AUC values for macro and micro, 0.93 and 0.97 for an overall classification (comparing 0.91 and 0.96 on the ResNet-50 model). To obtain better results using small models for mobile device use-cases, the authors proposed a system that evaluates the performance of earlier melanoma recognition system using the HAM10000 database and got an accuracy of 87.84\% using MobileNet~\cite{r23}. Yuan et al. ~\cite{r24} used a skin lesion segmentation technique on ISIC~\cite{r25} and PH2~\cite{r26} datasets. They built a deep neural network with 19 layers, where they formed their own loss function with Jaccard distance. Authors in~\cite{r27} used VGG16~\cite{r28} based CNN architecture, implementing transfer learning techniques with an accuracy of 78.6\% on ISIC datasets.

Due to a lack of labeled data and a class-imbalance problem, skin lesion classification using computer-aided diagnostic tools has become a more challenging task. To aid dermatologists in making more accurate diagnostic judgments, data augmentation techniques based on generative adversarial networks (GANs)~\cite{r29} were introduced for skin lesion classification. To bypass the issue of annotated data researchers proposed a method~\cite{r30} generating realistic synthetic skin lesion images applying Generative Adversarial Networks (GANs). The actual goal was to create fine-grained, high-resolution synthetic images of skin lesions. Authors in~\cite{r31} proposed an over-sampling method for finding the inter-class mapping between under and over-represented class samples in a bid to generate under-represented class samples using cycle-consistent generative adversarial networks~\cite{r32}. These synthetic images are then used as additional training data in the task of detecting melanoma. They show performance improvements for melanoma classification when using only additional synthetic 700 images compared to 9000 augmented images.

Although this burgeoning literature on skin lesion classification has provided many useful insights, several gaps remain between model architecture design and size. The majority of studies were based on very deep networks and many consisted of multiple stages. Even though these models are achieving state-of-the-art results, these are very expensive to train, that questions the ability of these models to be deployable in a clinical setting. Others were medical-based studies that took place in the clinical environment. Such approaches are time-consuming and conflict with the interest of patients in sharing their own data. In a clinical setting, the goal is to classify skin lesions accurately in the simplest way possible using less expensive and accessible devices such as smartphones or embedded devices~\cite{r33}. Since deep models and multi-stage approaches are considered somewhat infeasible to be directly deployable in a clinical setting, there is a pressing need to develop learning methods that require less expensive hardware and are equally or comparatively better than deep or multi-stage frameworks. We may train small models with fewer parameters. It'll have less accuracy, which means the model won’t be able to classify the melanoma accurately. To this end, we identified Knowledge Distillation (KD) could be a possible approach to address the aforementioned issues. KD was proposed by Hinton~\cite{r34} in 2015. The concept of KD is basically passing the learned representations from a complex and cumbersome model (e.g., ResNet-50) to a particularly small model (e.g., 3-layer CNN), or in simpler words, the concept is to train a smaller architecture using distilled knowledge which is taken from a pre-trained larger model which is also referred as a teacher model. 
 
In various applications of machine learning, such as object detection, acoustical modeling and natural language processing, KD was utilized successfully~\cite{r35}. Some studies are related to our work with KD dedicated to real-life needs: skin cancer classification task~\cite{r36}, smart autonomous vehicles~\cite{r37}, drowsiness detection~\cite{r38}, human activity recognition~\cite{r39}, and many more. The main idea was that teacher and student should process the exact same amount of image crop and augmented images. They found that these aggressive augmentations and very long training schedules are the key to making model compression via KD work well~\cite{r40} in practice. The study was conducted on five well-known natural image classification datasets: flowers102, pets, food101, sun397 and ILSVRC-2012 (i.e., ImageNet). This enables the robustness of distillation recipes to be verified in a wide range of real circumstances. They achieved good accuracy on different types of natural image datasets. Even though they study on KD implemented with a variety types of image datasets, these are limited to the natural image domain. 

The applications of KD are hardly explored in the medical domain and it remains an open question as to how KD translates in the medical domain. If KD can be efficiently utilized, it would tremendously reduce the complexity and compute expense involved in accurately identifying the correct class: benign or malignant. Further, given the differences involved between natural images and medical images, it is still a question of how KD would translate to medical image analysis tasks.

Authors~\cite{r1*} applied KD in image segmentation tasks with CT images; an approach was taken to fit this in lightweight systems. Mutual Knowledge Distillation (MKD) has been applied here ~\cite{r2*} to perform image segmentation tasks with the CT images from MRI data. Semi supervised learning is being implemented with KD and asymmetric label sharpening (ALS) algorithms~\cite{r3*} were initiated to imbalance problems in the chest X-RAY images and this attempt works fine in  large-scale annotated medical image data. Researchers worked on an `Ocular Disease' detection task where they proposed Multi-task Learning (MTL)~\cite{r4*} which takes the gist of KD and authors tried to find out best outcomes.

There also exists other model complexity reduction methods such as quantization~\cite{r46} and pruning~\cite{r45}. Quantization~\cite{r46} is the method of approximating a trained neural network with low bit width numbers, with the goal to reduce both the memory requirement and compute cost of using those models. Quantization is a method which is also applicable after training of deep neural nets via pruning or KD. Pruning~\cite{r45} is a technique by which a neural network is made smaller and efficient by eliminating the values of the weight tensors, with the goal to get a computationally cost-efficient model. It is an expensive process since it requires multiple iterations of pruning to get the final model. To reduce this compute cost, random pruning strategy~\cite{mittal2018recovering} is studied which is comparable to state of the art pruning methods. There is also Lottery Ticket Hypothesis (LTH)~\cite{r47} which essentially produces sparse networks. Note that this approach does not reduce the size of the model but rather makes some of the weights of the model to be zeros. Our work is significantly different in the sense that were are interested to rapidly building and deploying models in a clinical setting. Therefore, we find KD to be a viable approach towards this specific problem setting that reduces model complexity during testing on a single step which is crucial for model deployment. Previous studies have demonstrated this to be useful mostly in natural images but also few works in medical domain~\cite{r1*}~\cite{r2*}~\cite{r3*}~\cite{r34}~\cite{r35}~\cite{r37}~\cite{r40}.

\section{Method} \label{Method}
This section represents the main components, methodologies and algorithmic approaches of the proposed task of melanoma classification.

\subsection{Knowledge Distillation with Distill Student Network (DSNet)}
Despite the fact that neural models have been shown to be satisfactory in many disciplines, even those with complicated problem statements, the models are enormously big, with millions of parameters. This causes a barrier to these models being deployed on edge devices having low memory requirements. Knowledge Distillation (KD) is a possible solution to deal with this challenge.
\begin{figure*}[!ht]
\centering
\includegraphics[scale=.50]{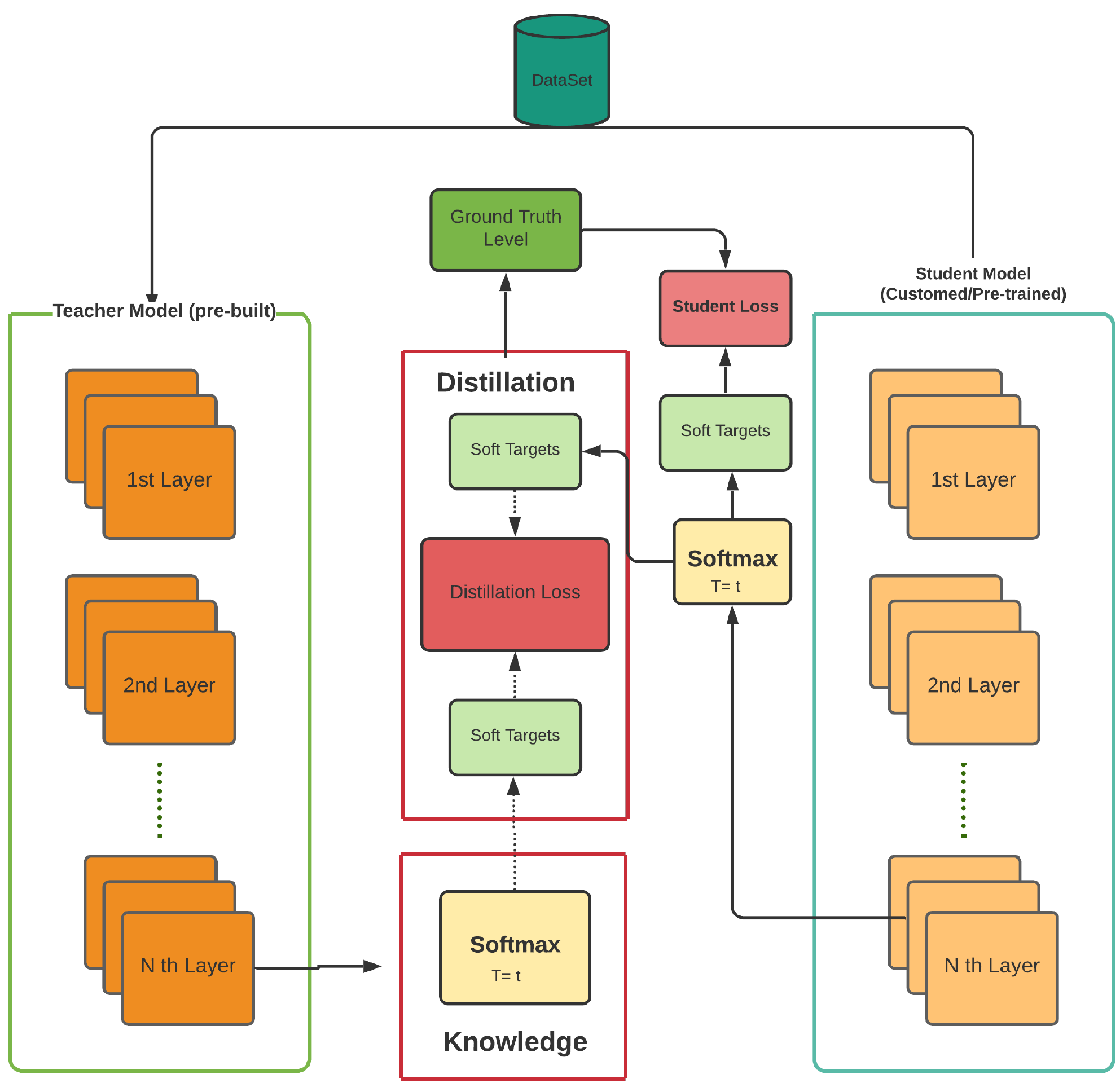}
\caption{Knowledge Distillation procedure in classification task, knowledge is being distilled from teacher architecture and being passed to the student model.}
\label{Fig:KDProcedure}
\end{figure*}

In Figure~\ref{Fig:KDProcedure}, we showed the overall skeleton of the whole KD approach as how it actually worked in our research. The concept behind KD is to compress heavy-weight models into light-weighted models with a significant tradeoff in accuracy. In KD, a small model (e.g. student) is trained using distilled knowledge from a pretrained larger model (e.g. teacher). For an image classification problem, the teacher model is trained on a image based dataset firstly and their labels as well. This pretrained teacher model outputs class probabilities based on an mage input. Also a DNN, but less complicated (i.e. has fewer layers and parameters), the student network may be trained on the same dataset. There are two components to the student model's training loss. The first minimizes the gap between the projected labels and the real labels (hard labels), while the second minimizes the teacher's predictions (i.e., soft labels). When the teacher model classifies the input data using soft labels (i.e., relative probabilities of distinct classes), this information is highly relevant. During this distillation process, temperature (T) plays an significant role in teaching the student model completely. The temperature helps to get a probability score. To get the probability score, first excerpts the logits from the teacher model and pass them through the softmax function with the selected temperature. By doing this, it gave softened probabilities for the student model.
The regular softmax function takes the logits and converts them into probabilities. Equation~(\ref{eq:regularsoftmaxwithoutT}) shows a softmax function and Equation~(\ref{eq:withTsoftmax}) show a softmax function after softening with temperature (T) which is defined as:
\begin{equation}
\begin{split}
\text{Softmax}(a_{i}) = \frac{\exp(a_i)}{\sum_{j=1}^{k} \exp(a_j)}
\end{split}
\label{eq:regularsoftmaxwithoutT}
\end{equation}
\begin{equation}
\begin{split}
\text{Softmax}(a_{i}) = \frac{\exp(a_i/T)}{\sum_{j=1}^{k} \exp(a_j/T)}
\end{split}
\label{eq:withTsoftmax}
\end{equation}

An image based dataset can be denoted as $\{X_i,Y_i \}$ $(X_i: images,\;Y_i: image\;labels)$. By passing an image $\ x \in X$ to the teacher and student, we get two sets of logits $a_t$ and $a_s$. We can define the probability distribution of teacher and student model such below:
\begin{equation}
\begin{split}
\text{P}(teacher) = softmax{(a_t/T)}
\end{split}
\label{eq:softmaxTeacher}
\end{equation}
\begin{equation}
\begin{split}
\text{P}(student) = softmax{(a_s/T)}
\end{split}
\label{eq:softmaxStudent}
\end{equation}

Here, T is determined as the distillation temperature which controls the P(teacher) and P(student) entropy. The distillation loss is defined as:
\begin{equation}
\begin{split}
loss=\alpha \;* \;L_{\text {hard }}(P({student}),Y)\\ +\;(1-\alpha)\;*\;L_{\text {soft}}(P(\text {student}),\;P(\text {teacher}))
\end{split}
\label{eq:loss}
\end{equation}

where $\alpha$ is the coefficient to weight the student and distillation loss, Y is the true label (i.e., hard label) for X (i.e., a one-hot vector), and $L_{hard}$, $L_{soft}$ is measured by cross-entropy. The entropy in P(student) proportionally increases with the value of T, which lead the Student to learn the relevant  probabilities of particular classes from the pretrained teacher model. Despite this, a large T can also increase the probability of irrelevant classes.

In this paper, we introduce the concept of knowledge distillation, which aims to distill and transfer knowledge between models in order to compress them. In traditional model compression applications~\cite{r41}, the student is usually required to share similar model nature (precisely, network architecture) with the teacher model. In our case, however, the purpose is to make the student as accurate as possible but not to completely mimic the behaviour of teacher architecture, because the student is much simpler.

In this research, we selected ‘ResNet50’ as the Teacher model, and then we built a model with convolutional nets as our Student model and named it \textbf{Distill Student Network (DSNet)}, and we took a different range of temperature (T) values to find out which T value gives the best outcome. After that, we tuned some pretrained architectures which have similar or greater parameters than our student ‘DSNet’ model to understand the scenario with different performance metrics for each and every architecture used as the ‘Student’ model, with the best T value found from the ‘DSNet’ model.

\medskip\noindent\textbf{ResNet-50 as Teacher.}\quad 
Deep neural networks have led to a number of image classification breakthroughs. Many other tasks of visual recognition have benefited greatly from very profound models. Over the years, therefore, there is a tendency to deepen and to solve more difficult tasks as well as to improve accuracy. However, as we go deeper, neural network training becomes harder and accuracy begins to saturate and also degrade. Such problems are being solved by residual learning.

The ResNet50 architecture is our primary model to be used as the teacher model for the reasons discussed above. ResNet indicates ‘Residual Network’~\cite{r16} that defines ‘residual learning’ as the key terminology introduced by this network. ResNet50 is a deep network of residuals with 48 convolutional layers, an Average Pooling layer and a fully connected (FC) layer, and this model is very popular for image classification tasks~\cite{r42}~\cite{r43}.

\begin{figure}[!htb]
\centering
\medskip
\includegraphics[scale=.30]{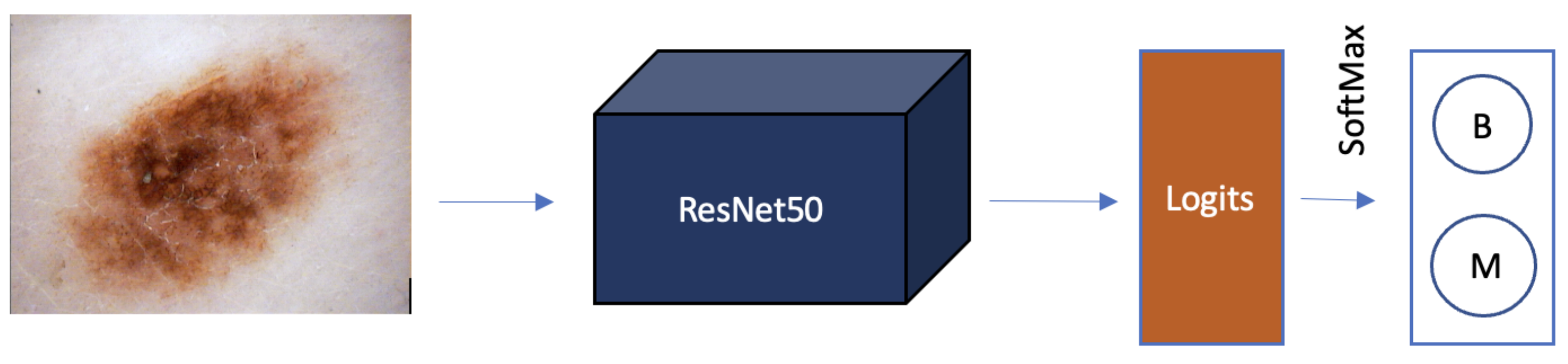}
\caption{Teacher Model, dermoscopic image is passed through this pretrained ResNet50 architecture and then moving forward to distillation process.}
\label{Fig:TResNet}
\end{figure}

In our model, we use the pretrained version of ResNet50 which has around 23,542,786 trainable parameters and the architecture in Figure~\ref{Fig:TResNet}. The pretrained model took weights from the ImageNet dataset. Our model had a 0.5 dropout and used the Softmax in the last layer which gave us the probability between the two classes. The network is optimized using Adam~\cite{r44} to minimize the categorical cross-entropy loss.

\begin{figure}[!ht]
\centering
\includegraphics[scale=.30]{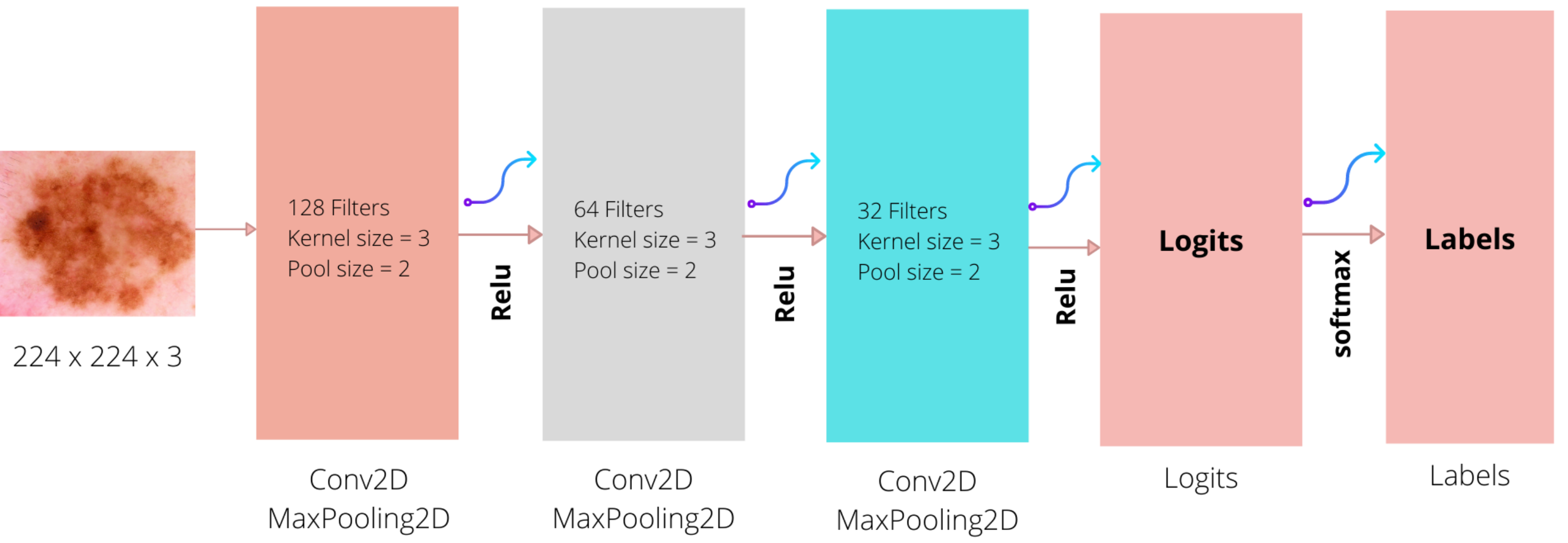}
\caption{Student Model, our built convolutional architecture that we named after ‘DSNet’.}
\label{Fig:DSNetArch}
\end{figure}

\medskip\noindent\textbf{DSNet as Student.}\quad 
In our work, we used a small architecture to build our student model, naming it the Distilled Student Network (DSNet). In the student model, we had only 268,898 trainable parameters which is much smaller than our teacher model. We used three layers of convolution and max pooling. We used a dropout of about 0.25 and lastly used the softmax function to get the probabilities, shown in Figure~\ref{Fig:DSNetArch}.

\subsection{Pretrained Student Models}
As we discussed earlier, we use pretrained models such as MobileNet, Inception V3. EfficientNet B0, VGG16, ResNet50, ResNet101 as the `Student' model, and applied the best Temperature value achieved from our `DSNet' model. Our goal is to understand the outcomes of different pretrained models as student and also to compare with our built `DSNet' model. In this section we will discuss these pretrained models.

\begin{figure}[!h]
\centering
\includegraphics[scale=.30]{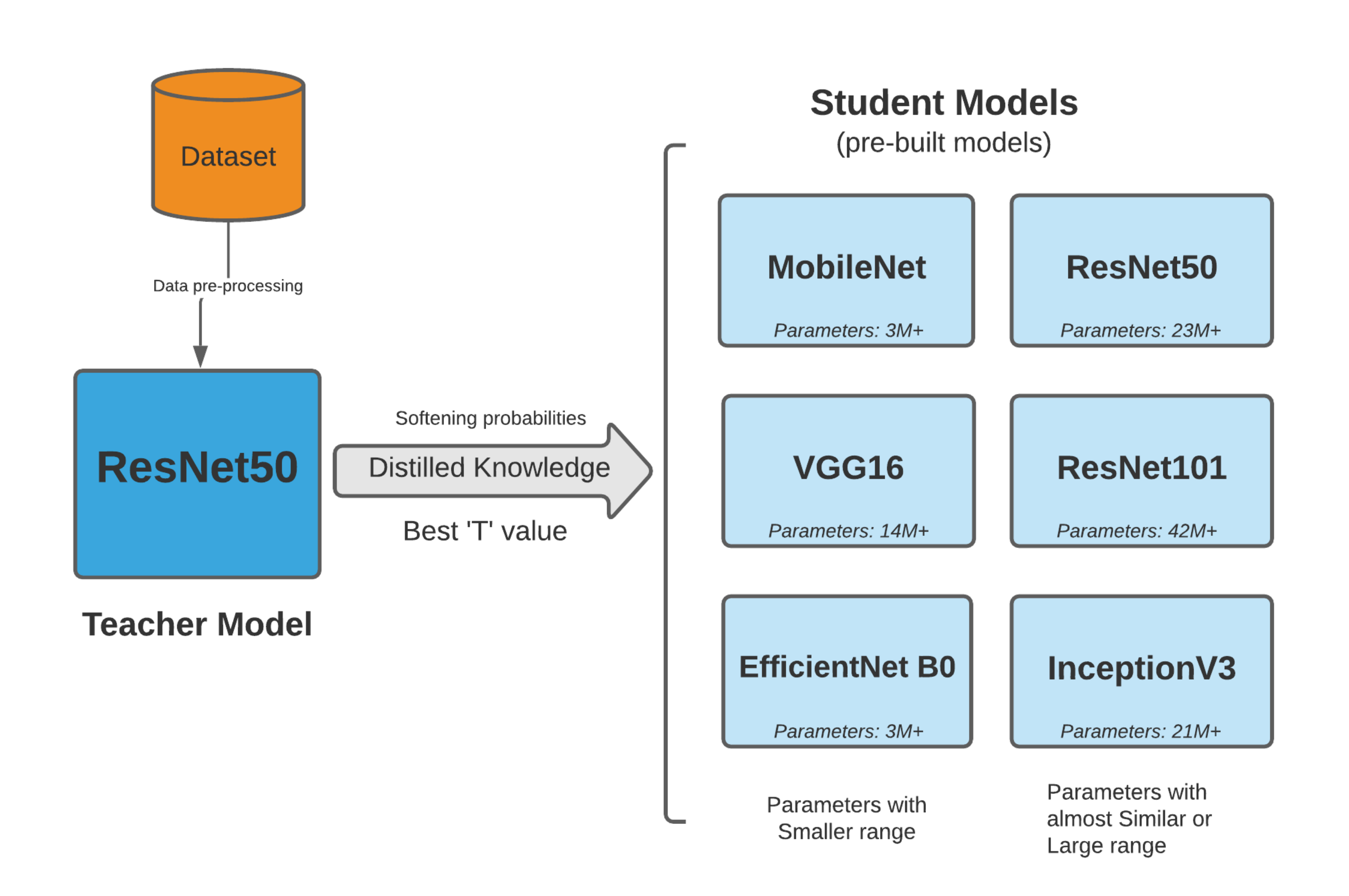}
\caption{Knowledge Distillation method with Pretrained models; Choose the pretrained model according to its parameters; MobileNet, EfficientNet B0, ResNet50, ResNet101, Inception V3, VGG16 all of these models are used also as student model for being compared with our ‘DSNet’.}
\label{Fig:PretrainedMod}
\end{figure}

In Figure~\ref{Fig:PretrainedMod}, we showed the workflow of pretrained student models. First we trained teacher model and extracted the logits and applied softmax with temperature to get the soften probabilities and then trained them to find the performances of these models as  student models. In this section we’ll discuss these pretrained models.

\medskip\noindent\textbf{ResNet-50 \& ResNet-101.}\quad 
In classification tests, Residual Networks (ResNet) are considered as one of the fastest performing deep neural networks. It won first place in ILSVRC's 2015 classification contest with an error margin of 3.57 percent in the top 5. A residual block is a stacking of layers in which the output of a single layer is taken and added to a layer further down in the block. ResNet's central concept is to introduce "identity shortcut connections" that avoid one or even more tiers in the convolution layer. The exact dimensions must be enforced directly in order to implement the "identity shortcut connection". As we said above, ResNet50 contains 23.5 million parameters and ResNet-101 contains 101 deep layers and more than 42 million trainable parameters.

\medskip\noindent\textbf{Inception V3.}\quad 
InceptionV3 is a CNN based architecture made up of a 42-layer deep neural network. The symmetric and asymmetric architectural components of the Inception-v3 model include convolutional layers, max pooling layers, dropouts, average pooling and FC layers. With less than 25 million parameters, the state-of-the-art of this model is 21.2 percent top-1 and 5.6 percent top-5 error. We checked performance metrics over the test set with 21M+ trainable parameters in knowledge distillation for this study.

\medskip\noindent\textbf{VGG-16.}\quad 
VGG16 is a CNN based model that concentrates on having 3x3 filters with a single stride and always uses the same padding and max pooling layer of 2x2 filters with a stride of 2 instead of having huge quantities of hyper-parameters. The convolution and max pooling layers are organized in quite a similar way throughout the architecture. The outcome part of the model consists of 2 FC layers and a softmax output layer. This VGG16 network has more than 138 million trainable parameters which is a fairly massive network. In this work, we took the pretrained VGG16 model as a student model which contains only 14M+ trainable parameters.

\medskip\noindent\textbf{EfficientNet-B0.}\quad 
EfficientNets are extremely efficient in terms of computation, and they also obtained a substantial performance on the ImageNet dataset, with an accuracy of 84.4 percent top-1. It became popular because of its scalability. In order for CNN to improve exactness, it needs to have a heedful balance between depth, width, and resolution. EfficientNet utilizes a complex coefficient to consistently scale architecture depth, width and resolution in a mannered way. This methodology was downsizing the total number of trainable parameters. This work used EfficientNet B0 as a student model with only 4M+ trainable parameters.

\medskip\noindent\textbf{MobileNet.}\quad 
The MobileNet is designed for mobile embedded vision applications. A lightweight network, MobileNet, decreases trainable parameters and computation by utilizing depthwise distinct convolution to the network. Simultaneously, MobileNet's classification accuracy on the ImageNet dataset scarcely drops by 1\%. In any case, the parameters and computational complexity of the MobileNet model should be diminished to be suitable for cell phones with limited memory. The model has fewer parameters and a decreased computational expense. In this work, MobileNet with 3M+ teachable parameters is considered a student model.

\subsection{Algorithm}
Algorithm~\ref{algori:KDalgoooo} demonstrates the pseudocode of the implemented KD approach. It summarizes the major algorithmic phases of this approach. The training set includes dermoscopic images of skin lesions and their associated class labels. The particular classes are combined together in the first phase (for example, binary classification has two groups), and each image is resized to $224\times 224\times 3$ pixels. Then, import the pretrained model ResNet50 which is the teacher architecture. Train the teacher model. After training, implement the distillation. First, extract the logits from the teacher model. Select temperature $T = 10$ and apply softmax function with the temperature and get the soften probabilities. Next, build a small model which is used as the student model (DSNet) and defines the distillation loss. Lastly, train the DSNet model all over the training dataset. Finally, we evaluate trained ResNet50 and DSNet model on the test set.

\begin{algorithm}
  \caption{Pseudocode of Knowledge Distillation}
  \label{algori:KDalgoooo}
  \begin{algorithmic}[1]
    \REQUIRE $\mathcal{D}=\{(\bm{x}_1,y_1),\dots,(\bm{x}_n,y_n)\}$:$\mathcal{D}$ determines the training set of dermoscopic images where the input is $\bm{x}_i$ and $y_i$ is a class label of .
    \ENSURE $\bm{\hat{y}}$ contains the predicted class labels.
    \FOR{$i=1$ to $n$}
    \STATE Assemble each lesion image in terms of classification label.
    \STATE Resize images by $224\times 224\times 3$.
    \ENDFOR
    \STATE import pretrained model ResNet50
    \STATE train teacher model(ResNet50) on data.
    \FOR{$i=1$ to $n$}
    \STATE training teacher
    \ENDFOR
    \STATE \textbf{Distill knowledge:}
    \STATE \quad Extract logit from teacher model
    \STATE \quad Select temperature T
    \STATE \quad Apply softmax with T and get soften probabilities
    \STATE Build DSNet (student) model
    \STATE define distillation loss
    \FOR{$i=1$ to $n$}
    \STATE training student
    \ENDFOR
    \STATE Model evaluation on the test set
  \end{algorithmic}
\end{algorithm}

\section{Experiments} \label{experiments}
In this section, we conducted adequate experiments to assess the performance of the knowledge distillation approach on the task of melanoma classification. The source code to reproduce the experimental results will be made publicly available on GitHub\footnote{https://github.com/Shakib-IO/KD-lesions}.

\subsection{Experimental Setup}
\begin{figure}[!htb]
\centering
\includegraphics[scale=.28]{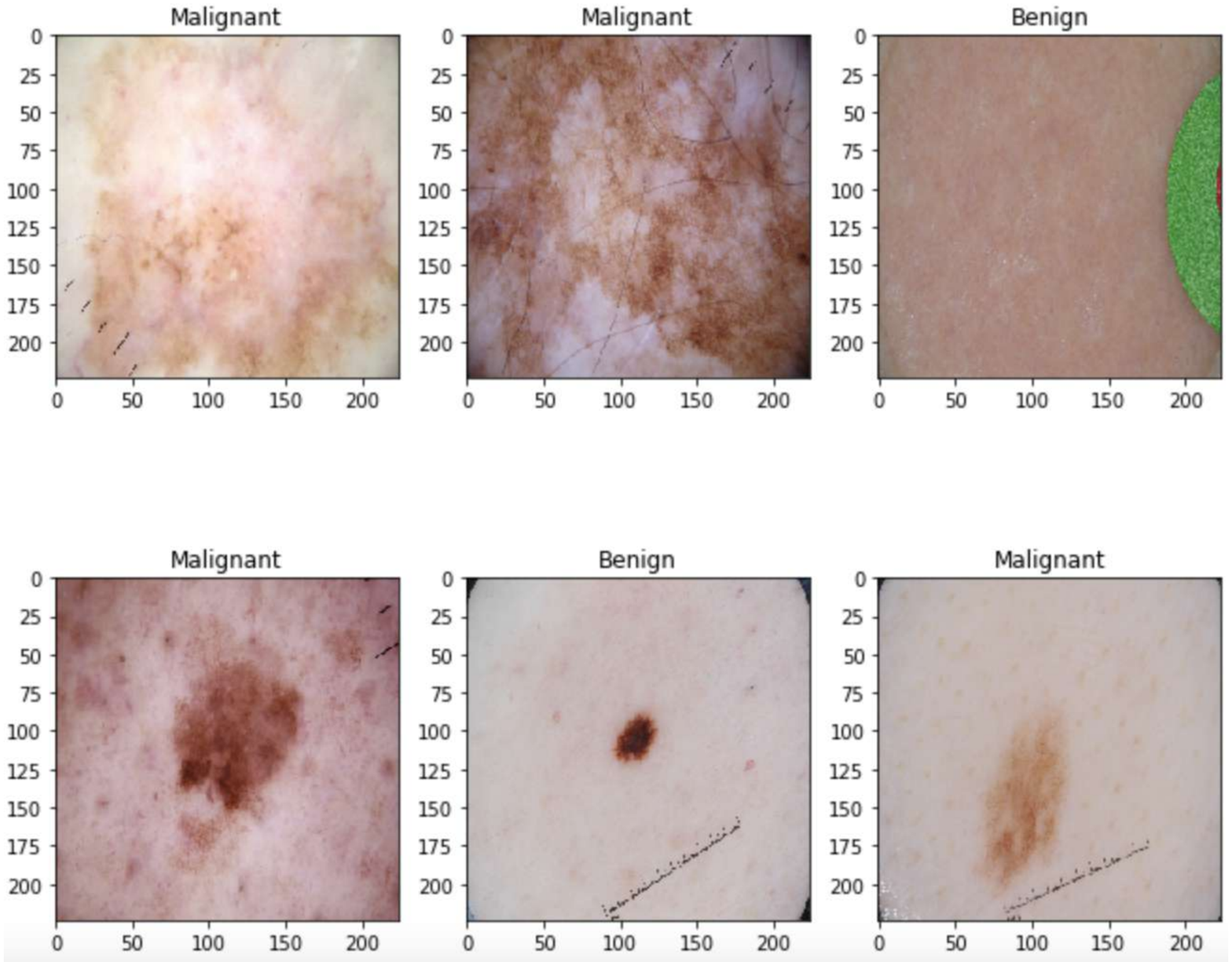}
\caption{Random input images from ISIC dataset with label of Benign or Malignant.}
\label{Fig:ISIC}
\end{figure}

\medskip\noindent\textbf{Dataset details.}\quad 
Every year, the International Symposium on Biomedical Imaging (ISBI) hosts challenges in different aspects of biomedical fields. The goal of the challenge was to provide a fixed dataset snapshot to support the development of automated melanoma diagnosis algorithms across 3 tasks of lesion analysis: segmentation, dermoscopic feature detection, and classification. One of the challenges is Skin Cancer detection. Hence, the viability of knowledge distillation is assessed on the ISIC dataset, an open access dermescopic image analysis benchmark challenge for melanoma classification using annotated skin lesion photos from the International Skin Imaging Collaboration (ISIC) database~\cite{r19}. These images include a variety of textures in the adjacent pixels, as well as low contrast, making melanoma classify a difficult task. To tackle this task, the dataset includes a blend of images of both malignant and benign skin lesions that were arbitrarily divided into training and test sets, with 6996 images in the training set and 3000 images in the test set, respectively. To train our model, we used 3500 images of benign and 3496 images of malignant and 1500 images for testing each class. Due to the multiple sizes of the images in the ISIC dataset, they were resized to $224\times 224\times 3$ pixels. Figure~\ref{Fig:ISIC} shows some samples from the dataset.

\medskip\noindent\textbf{Implementation Details.}\quad 
For this work, all experiments were run on Google Colab. Colab provides its own GPU. The Colab has 12 GB of RAM and an NVIDIA Tesla K80 GPU. To train all of the models, we used the Adam optimizer and a learning rate of 0.001. Categorical cross-entropy is used as a loss function. The batch size for training is 64~\cite{r7*}.

\medskip\noindent\textbf{Evaluation Metrics.}\quad 
The success of the knowledge distillation approach is determined by a thorough comparison of many performance evaluation metrics. The metrics included the Precision, Recall, F1 Score, Receiver Operating Characteristic (ROC) curve and the Area Under the ROC Curve (AUC). To calculate all of the metrics, need to find the TP (True Positive), TN (True Negative), FP (False Positive), FN (False Negative) values. True Positive indicates that the output of the real class is yes, and the output of the predicted class is also yes, whereas True Negative indicates that the value of the real class is no, and the value of the anticipated class is no. False Positive indicates that the real class is no while the predicted class is yes, whereas a False Negative indicates that the real class is yes but the expected class is no. In this work, the TP is the number of correctly predicted malignant classes where the actual class is malignant, while the number of correctly predicted benign lesions is denoted by TN. At the same time, FP tells us that the number of wrongly predicted malignant classes where there are none exists, whereas FN is the number of predicted benign lesions but the actual class is malignant.

The accuracy metrics is the ratio of correct predictions  over the total number of predictions evaluated

\begin{equation}
\text{Accuracy} = \frac{\text{TP}+\text{TN}}{\:\text{TP}+\text{FP}+\text{TN}+\text{FN}\:}
\label{eq:Accuracy}
\end{equation}

Precision is defined as the proportion of correctly predicted positive observations to total expected positive observations. Using this equation, calculation of the precision is

\begin{equation}
\text{Precision} = \frac{\text{TP}}{\:\text{TP}+\text{FP}\:}
\label{eq:Precision}
\end{equation}

Recall or true positive rate (TPR) tells the proportion of actual positive samples that got predicted as positive. To get the value recall using this equation

\begin{equation}
\text{Recall} = \frac{\text{TP}}{\:\text{TP}+\text{FN}\:}
\label{eq:Recall}
\end{equation}

The average of Precision and Recall is denoted as F1 Score. Here is the equation given below

\begin{equation}
\text{F1 Score} = 2\times\frac{\text{Precision}\times\text{Recall}}{\:\text{Precision}+\text{Recall}\:}
\label{eq:F1Score}
\end{equation}

Another metric, AUC, indicates the information in the ROC curve, which shows TPR versus $\text{FPR}=\text{FP}/(\text{FP}+\text{TN})$, the false positive rate, at variety of thresholds. Larger AUC values determine better performance at differentiating malignant or benign in terms of melanoma classification.

\medskip\noindent\textbf{Varying the Distillation Temperature.}\quad 
As discussed earlier, the logits we get from the teacher model pass through the softmax function. To get softened probabilities for the student model, the logits are divided by the temperature. Temperature played an integral role in teaching the student model thoroughly. In this work~\cite{r34},  they tried different ranges of temperatures in image domains. They concluded that all temperatures above 8 produced similar softening probabilities, while temperatures in the range of 2.5 to 4 produced much better results than temperatures higher or lower. However, as this work involved medical images, we tried to find out which temperature is most suitable to make the soften probability. So for this, we applied various temperatures where we wanted to see how the values changed with temperature values. This motivated us to try different ranges of temperatures $T = [1,3,5,7,10,20,50,70,90,100]$. We applied all these temperatures and used the DSNet student model to identify the most suitable configuration (i.e. which value of T gives a high accuracy score).

\begin{figure}[!htb]
\centering
\includegraphics[scale=.30]{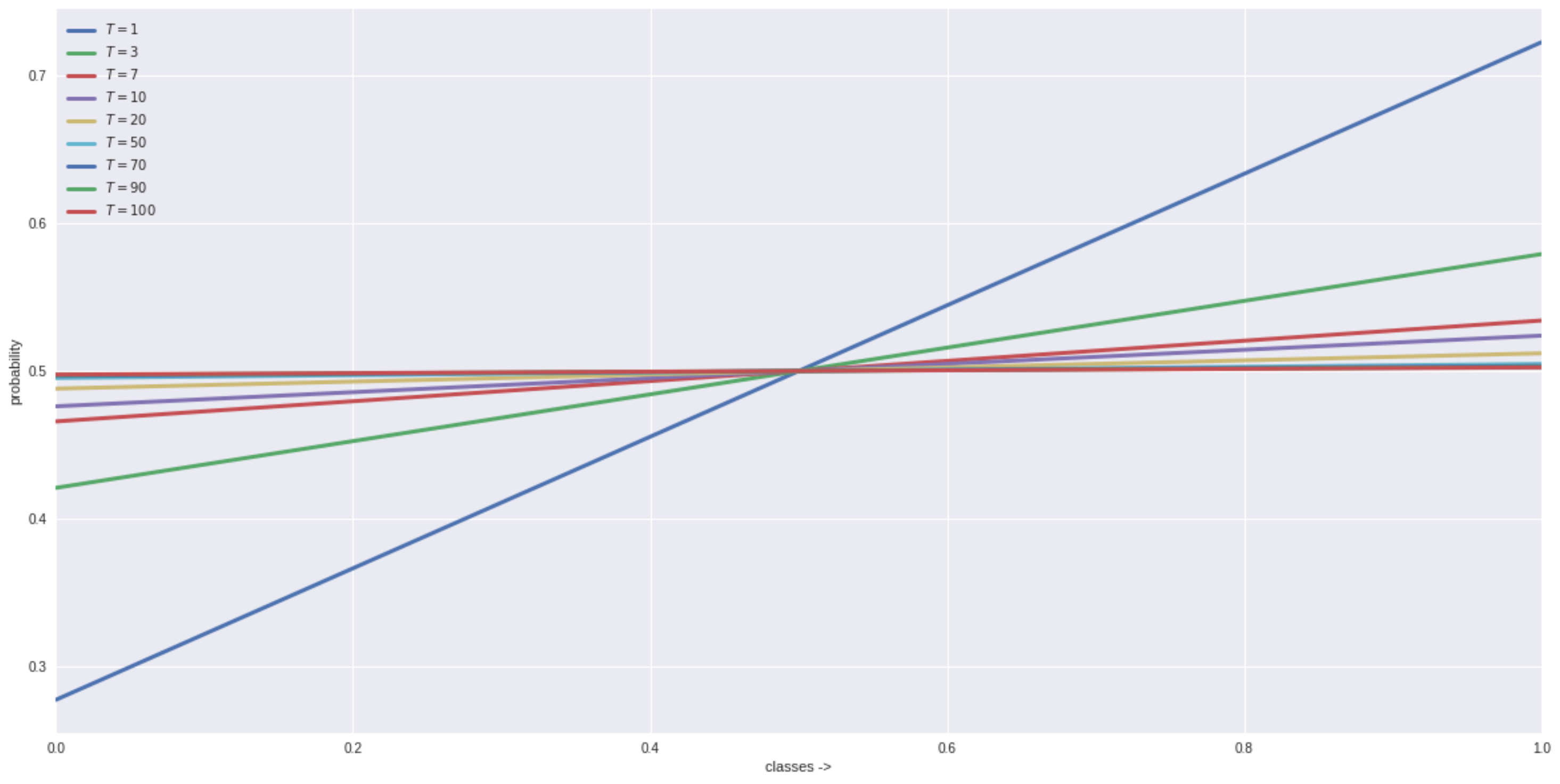}
\caption{Probability Distribution for different temperature (T) values; different T value gives different probability score.}
\label{Fig:ProbabilityDistriofT}
\end{figure}

Figure~\ref{Fig:ProbabilityDistriofT} indicates how the temperature changes the probability distribution over the binary class. When the temperature is in the initial stage and apply softmax with it, the probability score is more likely to have a high difference between classes. As we increase the temperature value the probability score is going to balance between two classes which are more likely to have better knowledge.
From Figure~\ref{Fig:ProbabilityDistriofT} we can see that, when we applied $\text{T}=\text{3}$, the probability score was around 0.41 and 0.59, simultaneously when  $\text{T}=\text{10}$, the probability score we got was around 0.5 and 0.5. So, for T=10 we found the probability is balanced equally between two classes, which helps the model become less biased. 

\subsection{Results and Analysis}
In this section, we demonstrate the performance of the Teacher model and then Student models that are directed by the distill knowledge of the teacher model.

After training the teacher model with ResNet50, we found an accuracy of 99.6\%. The model took 596 seconds to run all the epochs and to predict the outcome it took 10.25 seconds.

After that, we found the best value for temperature (T) using our DSNet model. We applied this T value to different pretrained models to see how the accuracy and time changes, so that we could compare them with our proposed DSNet and come up with an analysis.

\medskip\noindent\textbf{Verifying accuracy for different T values on DSNet.}\quad The student model will take the logit from the trained teacher model and will soften it by dividing it by the temperature. To find out which temperature the model can predict more accurately, we apply the DSNet model which has 0.26 million parameters. Furthermore, We chose these T values because we observed that the outcomes fluctuated when we increase the temperature value. Therefore, we raise the temperature value frequently. Temperature values $T= 3, 5, 7, 10, 20, 50, 70, 90, 100$ are applied to the softmax layer and we checked the accuracy after training with the same factors as Teacher, but 20 epochs, shown in Table~\ref{Tab:Temperature}.
\begin{table}[!ht]
\caption{Accuracy for different temperature T values on DSNet, intends to check the best T value to apply further.}
\centering
\medskip
\scalebox{0.85}{
\begin{tabular}{c c |c| c}
\toprule
Student Model & Trainable Parameters & Temperature(T) & Accuracy\\
\midrule
\multirow{8}{*}{DSNet} & \multirow{8}{*}{0.26 Million} & 3 & 0.828\\
        & & 5 & 0.850\\
        & & 7 & 0.896\\
        & & 10 & 0.917\\
        & & 20 & 0.842\\
        & & 50 & 0.813\\
        & & 70 & 0.898\\
        & & 90 & 0.866\\
        & & 100 & 0.772\\
\bottomrule
\label{Tab:Temperature}
\end{tabular}}
\end{table}

From Table~\ref{Tab:Temperature}, we can see how accuracy changes with different temperature values on our proposed DSNet model. And here we can clearly observe that $\text{T}=\text{10}$ gives a better accuracy of 91.7\% which is higher than the other ranges of temperature. So, here we find the temperature value $\text{T}=\text{10}$ works well in our work and we tried different pretrained models to check how training \& prediction time changes with this temperature and how the accuracy varies as well.

\medskip\noindent\textbf{Student model performance.}\quad
Pretrained student models are - MobileNet, VGG16, Inception V3, EfficientNetB0, ResNet50, ResNet101.These models contain individual parameters and we checked how much time they took to train and predict the results. Then we also checked  the accuracy which is the key part as which model can predict the malignancy more accurately and sooner. Table~\ref{Tab:studentmodeloutcomes} shows the overall outcomes of student models.
\begin{table*}[!tb]
\rowcolors{1}{}{lightblue}
\caption{Student model outcomes, pretrained models and DSNet is being compared in terms of Parameters, Time and Accuracy.}
\medskip
\centering
\scalebox{1}{
\begin{tabular}{l l l l c}
\toprule
Student Models & Trainable Parameters & Training Time & Prediction Time & Accuracy\\ 
\midrule
ResNet101  & 42.5 Million & 1972 seconds & 14.55 seconds & 0.996 \\
ResNet50 & 23.5 Million & 1165 seconds & 8.28 seconds & 0.994 \\
InceptionV3 & 21.7 Million & 896 seconds & 10.26 seconds & 0.995 \\
VGG16 & 14.7 Million & 664 seconds & 20.49 seconds & 0.998 \\
EfficientNetB0 & 4 Million & 1070 seconds & 4.98 seconds & 0.820 \\
MobileNet & 3.2 Million & 660 seconds & 5.14 seconds & 0.985 \\
\textbf{DSNet} (\textit{ours}) ~ & \textbf{0.26 Million} & \textbf{340 seconds} &\textbf{2.57 seconds} & \textbf{0.917} \\
\bottomrule
\end{tabular}
\label{Tab:studentmodeloutcomes}}
\end{table*}

From Table~\ref{Tab:studentmodeloutcomes}, we can see the changes in Training \& Prediction time and accuracy vary with different pretrained models of different parameters. Our motivation is to verify and understand the change in accuracy and time of pretrained models which have similar, lesser and greater parameters than the teacher model and to compare them with our built DSNet.

MobileNet, VGG16, Inception V3, EfficientNet B0 these models contain fewer parameters than the Teacher Model (ResNet50). ResNet50 by itself is also used here as a Student model to check how it works with the same teacher and student model. And ResNet101 is a bigger architecture than the Teacher model with almost 2 times larger parameters.

We compare the Student models, MobileNet and EfficientNet B0 have 3.2 million and 4 million parameters where MobileNet took 660 seconds to train the data and 5.14 seconds to predict, and EfficientNet B0 took 1070 seconds to train (comparatively higher than MobileNet) but 4.98 seconds to predict which is less than MobileNet. But the accuracy of MobileNet is 98.5\% which is much higher than the accuracy of EfficientNet B0 that is 82\%. Then if we move on to the bigger models, VGG16 and Inception V3 contain 14.7 million and 21.7 million parameters respectively. To train the model, VGG16 took 664 seconds and Inception V3 took 896 seconds which is fairly considerable due to the difference in the number of parameters. In the context of prediction time, VGG16 predicts the outcome in 20.49 seconds where Inception V3 predicts it in 10.26 seconds, that indicates that Inception V3 predicts faster than VGG16. The accuracy of VGG16 and Inception V3 is 99.80\% and 98.50\% respectively; that suggests that both models can predict almost accurately where Inception V3 predicts more early. ResNet50 is the model we chose for our teacher model. It is used as a Student model as well and it took 1165 seconds to train and 8.28 seconds to predict with 99.40\% accuracy. The much bigger model ResNet101 with 42.5 million parameters, took 1972 seconds to train and 14.55 seconds to predict and gave 99.60\% accuracy which is higher than the other pretrained models, but also took a huge amount of time compared to those models.

Our built DSNet model has a very inferior number of parameters like 0.26 million which is more than 10 times less than the MobileNet and undoubtedly less than the other models shown in Table~\ref{Tab:studentmodeloutcomes}. The training time of our model is \textbf{340} seconds which is less than any of the models above and the prediction time is \textbf{2.57} seconds, which means that our model can predict the malignancy much earlier than these pretrained models. Hence, our main goal was to develop a lightweight model that can be easily deployed on any small edge device, so with such substantial accuracy it is a good model to use on a small device. In terms of accuracy, On the test set, the DSNet model had an accuracy of \textbf{91.73\%}, which is impressive because it used fewer parameters and took less time to train and predict. However as we said earlier, our motivation is to make a model that is deployable on lightweight devices. So the number of parameters, as well as training \& prediction time is an important matter to consider with even slightly less accuracy. DSNet can be deployed to devices that don’t need that much high computational amenities. Obviously, more research on DSNet can improve its accuracy and other parameters as well.

\begin{table}[!h]
\caption{Performance Metrics of Teacher and Student model on Benign and Malignant Class.}
\medskip
\centering
\scalebox{0.63}{
\begin{tabular}{l |c c c |c c c}
\toprule
Classification & \multicolumn{3}{c|}{Malignant (Class 1)} & \multicolumn{3}{c}{Benign (Class 0))}\\
\midrule
Metrics & Precision & Recall & F1 Score & Precision & Recall & F1 Score\\
\midrule
Teacher Model (ResNet50) & 1.00 & 0.99 & 1.00 & 0.99 & 1.00 & 1.00\\
Student Model (DSNet) & 0.96 & 0.87 & 0.91 & 0.88 & 0.97 & 0.92\\
\bottomrule
\end{tabular}}
\label{Tab:MetricsofTechStumodels}
\end{table}

Here Table~\ref{Tab:MetricsofTechStumodels} shows the outcomes of performance metrics that we discussed above. Teacher model shows that it can predict the malignant class correctly as it’s recall value is 0.99 and the predicted outcome is totally correct because precision gives a value 1.00, and the F1 score of 1.00 also validates that both predicted outcomes are true. The same goes for the benign class for teacher model as the F1 score gives a value of 1.00, so undoubtedly, the teacher model’s prediction can be said to be accurate. DSNet as our student model gives a recall value of 0.87 for malignant and the  value of benign is 0.97, which can help to classify more perfectly that the lesion is malignant or not. The prediction of the malignant class is also be called almost correct as precision gives a value of 0.96. As our DSNet contains very few parameters, the classification varies with that but it can still be a helpful model and there are chances to improve it with more research we believe.

To measure the performance of other pretrained models as student models, MobileNet, Inception V3, VGG16, ResNet50, ResNet101 performed almost exactly in classifying malignancy where EficientNet B0 showed comparatively less performing in malignant class compared to others. We can also see that VGG16 and ResNet101 performs even better than the teacher model to classify malignancy, shown in Table~\ref{Tab:PerformanceMetricsofStudentmodels}.

\begin{table}[!ht]
\caption{Performance Metrics of Pretrained Student model on Benign and Malignant Class.}
\medskip
\centering
\scalebox{0.57}{
\begin{tabular}{l |c c| c c| c c}
\toprule
Metrics & \multicolumn{2}{c|}{Precision} & \multicolumn{2}{c|}{Recall} & \multicolumn{2}{c}{F1 score}\\
\midrule
Classification & Benign (0) & Malignant (1) & Benign (0) & Malignant (1) & Benign(0) & Malignant (1)\\
\midrule
MobileNet & 0.98 & 0.99 & 0.99 & 0.98 & 0.99 & 0.99\\
Inception V3 & 0.99 & 0.98 & 0.98 & 0.99 & 0.98 & 0.99\\
EfficientNet B0 & 0.80 & 0.85 & 0.86 & 0.78 & 0.83 & 0.81\\
VGG16 & 1.00 & 1.00 & 1.00 & 1.00 & 1.00 & 1.00\\
ResNet50 & 0.99 & 1.00 & 1.00 & 0.99 & 0.99 & 0.99\\
ResNet101 & 1.00 & 0.99 & 0.99 & 1.00 & 1.00 & 1.00\\
\bottomrule
\end{tabular}}
\label{Tab:PerformanceMetricsofStudentmodels}
\end{table}

These performance metrics in Table~\ref{Tab:PerformanceMetricsofStudentmodels} show how these student models performed in malignant classification. The performance of EfficientNet B0 was lower than other models. If we look at Table~\ref{Tab:Comparison}, the EfficientNet B0 with 4 million parameters gave less accuracy, precision, recall and f1 value than the 15 times smaller model DSNet which contains only 0.26 million parameters. That indicates, DSNet performs even better than EfficientNet B0 in this Knowledge Distillation approach to Melanoma classification.

\begin{table*}[!ht]
\rowcolors{0}{}{aqua}
\caption{Comparison of EfficentNetB0 and DSNet on performance metrics, DSNet performed better in Melanoma classification.}
\medskip
\centering
\scalebox{0.95}{
\begin{tabular}{l |c c c |c c c}
\toprule
Classification & \multicolumn{3}{c|}{Malignant (Class 1)} & \multicolumn{3}{c}{Benign (Class 0))}\\
\midrule
Metrics & Precision & Recall & F1 Score & Precision & Recall & F1 Score\\
\midrule
EfficientNetB0 (4 Million parameters) & 0.80 & 0.85 & 0.86 & 0.78 & 0.83 & 0.81\\
DSNet (0.26 Million parameters) & 0.96 & 0.87 & 0.91 & 0.88 & 0.97 & 0.92\\
\bottomrule
\end{tabular}}
\label{Tab:Comparison}
\end{table*}

We evaluated the AUC for the teacher model (ResNet50) and the student model (DSNet). We observed in the teacher model, achieving an AUC of 99\% for the benign class and 98\% for the malignant class. AUC of 95\% for both malignant and benign in the student model. 
\begin{figure}[!htb]
\centering
\includegraphics[scale=.35]{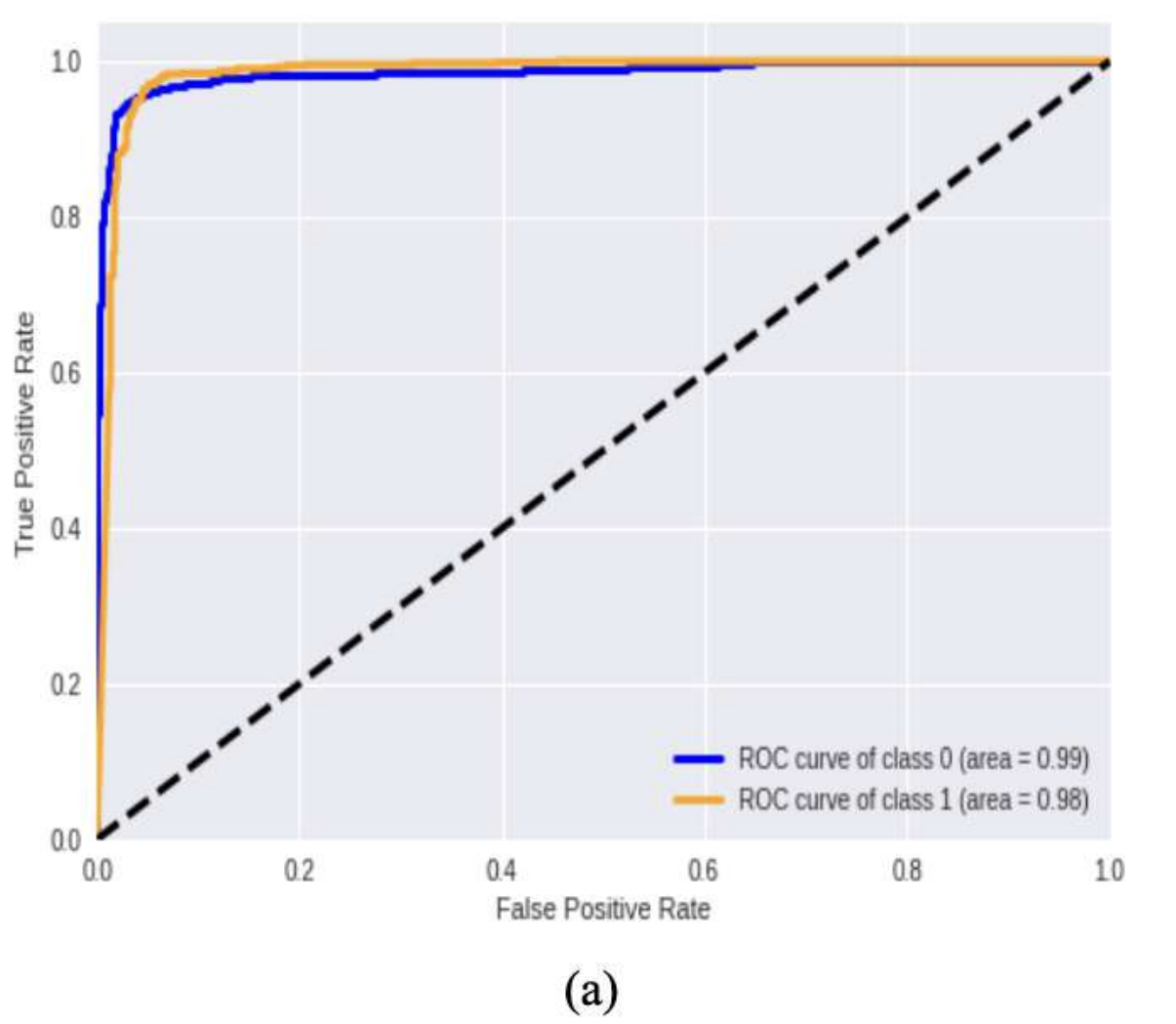}
\includegraphics[scale=.35]{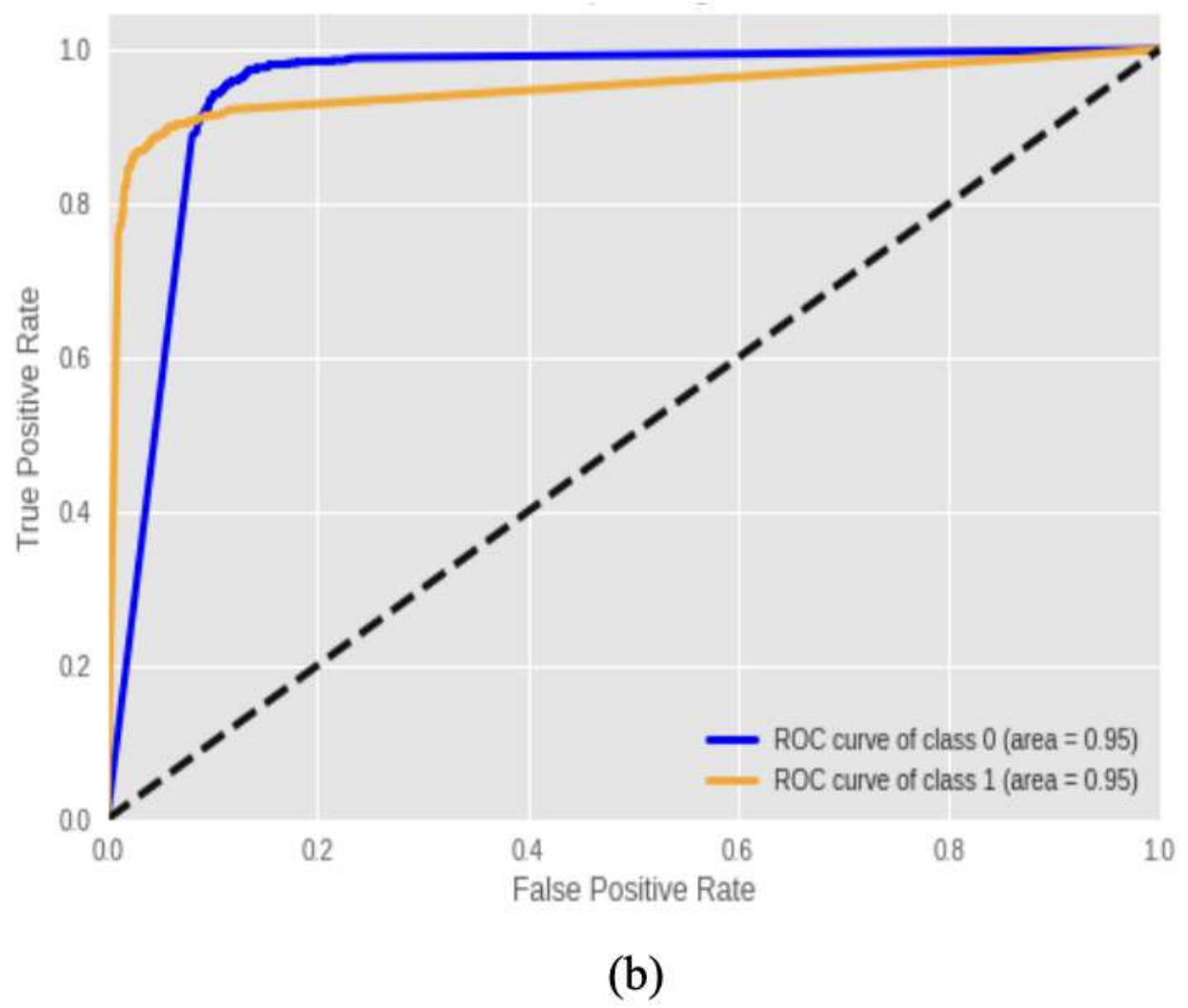}
\caption{ROC curves (a) for Teacher Model (ResNet50) and (b) for Student Model (DSNet) along with the corresponding AUC values.}
\label{Fig:AUC}
\end{figure}
Figure~\ref{Fig:AUC} displays the ROC curve, which depicts our teacher model and the student model performance are relatively close. Each factor on ROC is a distinction between false positives and false negatives. A better performance is indicated by a ROC curve that is near the upper right. In the teacher model ROC curve, it has most parts in the upper right which means its true positive rate is much higher that means it has better performance. Whereas the student model took knowledge from the teacher and then tried to predict classes. So, in the student model ROC curve, we can see that the true positive rate is quite higher and very relative to the teacher model which indicates our student model performance is considerably better.

\begin{figure}[!ht]
\centering
\includegraphics[scale=.25]{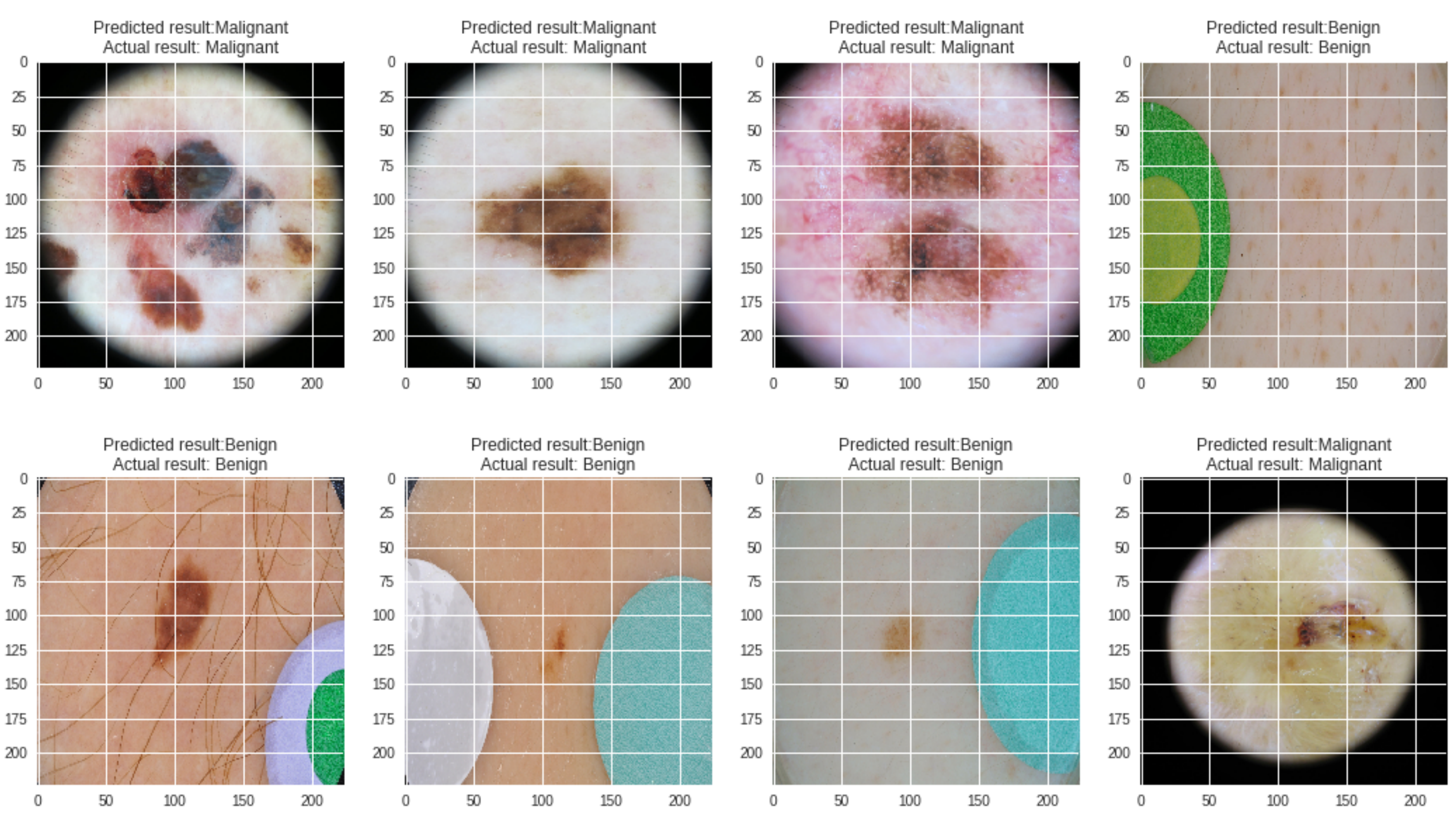}
\caption{Prediction of Malignant and Benign classes with DSNet model.}
\label{Fig:classifyOfMelaNoma}
\end{figure}

After training the student model with distilled knowledge from the teacher, we checked if it could easily identify skin lesion images as malignant or benign, shown in Figure~\ref{Fig:classifyOfMelaNoma}.
Here, we took random images from the test set and tried to see how our DSNet could predict Malignant or Benign. All the pictures showed that the DSNet correctly identified which dermoscopic image was benign or malignant. With such a small number of parameters, our student model can predict as accurately as we expected and this is clearly understandable, so the DSNet performed appropriately in detecting melanoma.
Our finding states that knowledge distillation is an appropriate choice for medical image analysis where a small student network can easily identify the true class. Because DSNet has such a small parameter and requires little computation power to run, it was easily deployed on edge devices since it takes less time to predict and train with good accuracy. 

\medskip\noindent\textbf{Parameter sensitivity analysis.}\quad We experiment with different batch sizes and find that the accuracy of the model at different batch sizes are relatively close to each other. For batch sizes greater than 96 (i.e. 128, 256), GPU memory gets full and crashes. Table ~\ref{Tab:Parameteranalysis} shows that our proposed model is less sensitive to changes in hyperparameters.

\begin{table}[!ht]
\caption{Parameter Sensitivity Analysis.}
\medskip
\centering
\begin{tabular}{c |c}
\toprule
BATCH SIZE & ACCURACY OF DSNet (\%)\\
\midrule
18 & 0.881\\
32 & 0.897\\
64 & 0.917\\
96 & 0.831\\
\bottomrule
\end{tabular}
\label{Tab:Parameteranalysis}
\end{table}

\section{Conclusion} \label{conclusion}
In this paper, we propose a knowledge distillation approach in detecting melanoma from dermoscopic images. To distill knowledge, we use two models which are the teacher and the student model, ResNet-50 and DSNet respectively. At first, we use ResNet-50 as our teacher model with 23 million trainable parameters which is pretrained on ImageNet data. The teacher model achieved an accuracy of 99.60\% and took 10.25 seconds to predict the Malignant or Benign classes. To train the student model using the teacher model, we divided the probabilities by temperature and got the softened probabilities for the student model. We experimented with different temperature values and chose the right temperature for the student model which achieved best performance. We built the student model with basic convolution blocks and named it DSNet with 0.26 million parameters, which achieved 91.7\% accuracy and 2.57 seconds to predict the two classes. We also compared different pretrained CNN models with our DSNet to clarify how they perform as student models in the Knowledge Distillation method and found better results in inference runtime. Interestingly, we find that DSNet works better than the larger model EfficientNet-B0 in terms of both runtimes. Besides DSNet is a lightweight model which can be easily implemented in any clinical setting or memory constraint device. Reducing computational complexity is important and we find KD to be a potential approach to address this issue. One direction is to further improve the detection performance and close the gap between small and large models which we intend to explore in future work. We aim to explore pruning and quantization methods in more detail for the task of melanoma detection, with the goal to further reduce model complexity and training time.

\section*{\fontsize{13}{13}\selectfont Compliance with Ethical Standards}
\medskip\noindent\textbf{Funding Statement:}\quad
The authors received no funding from other sources; this is a self-funded research, and all associated expenditures were covered solely by the authors of this article.

\bibliographystyle{ieeetr}
\bibliography{paper}
\end{document}